\documentclass[10pt,twocolumn,letterpaper]{article}

 \usepackage{cvpr}              

\usepackage[dvipsnames]{xcolor}
\usepackage{ mathrsfs }
\usepackage{tikz}
\usetikzlibrary{arrows.meta, calc, decorations.pathreplacing}
\usepackage{pgfplots}
\pgfplotsset{compat=1.18}
\usepackage{tikzscale} 
\usepackage{import}     
\usepackage{threeparttable} 
\usepackage{algorithm}
\usepackage{algpseudocode}

\definecolor{cvprblue}{rgb}{0.21,0.49,0.74}
\usepackage[pagebackref,breaklinks,colorlinks,citecolor=cvprblue]{hyperref}

\def\paperID{281} 
\def\confName{3DV\xspace}
\def\confYear{2026\xspace}

\title{Pixel-Accurate Epipolar Guided Matching}

\author{Oleksii Nasypanyi$^{1}$, Francois Rameau$^{2}$\\
~$^{1}$Stony Brook University,USA~~~~$^{2}$SUNY Korea, Incheon, Korea \\
{\tt\small oleksii.nasypanyi@stonybrook.edu~~~~~~{\tt\small francois.rameau@sunykorea.ac.kr}}
}

\begin{document}
\maketitle
\begin{abstract}
Keypoint matching can be slow and unreliable in challenging conditions such as repetitive textures or wide-baseline views. In such cases, known geometric relations (e.g., the fundamental matrix) can be used to restrict potential correspondences to a narrow epipolar envelope, thereby reducing the search space and improving robustness.
These epipolar-guided matching approaches have proved effective in tasks such as SfM; however, most rely on coarse spatial binning, which introduces approximation errors, requires costly post-processing, and may miss valid correspondences.
We address these limitations with an exact formulation that performs candidate selection directly in angular space. In our approach, each keypoint is assigned a tolerance circle which, when viewed from the epipole, defines an angular interval. Matching then becomes a 1D angular interval query, solved efficiently in logarithmic time with a segment tree. This guarantees pixel-level tolerance, supports per-keypoint control, and removes unnecessary descriptor comparisons. 
Extensive evaluation on ETH3D demonstrates noticeable speedups over existing approaches while recovering exact correspondence sets.
%
The project page is available \href{https://lexanagibator228.github.io/Pixel-Accurate-Epipolar-Guided-Matching}{ here}.
\end{abstract}
    
\section{Introduction}
\label{sec:intro}

\begin{figure*}[tb]
    \centering
    \includegraphics[width=0.9\linewidth]{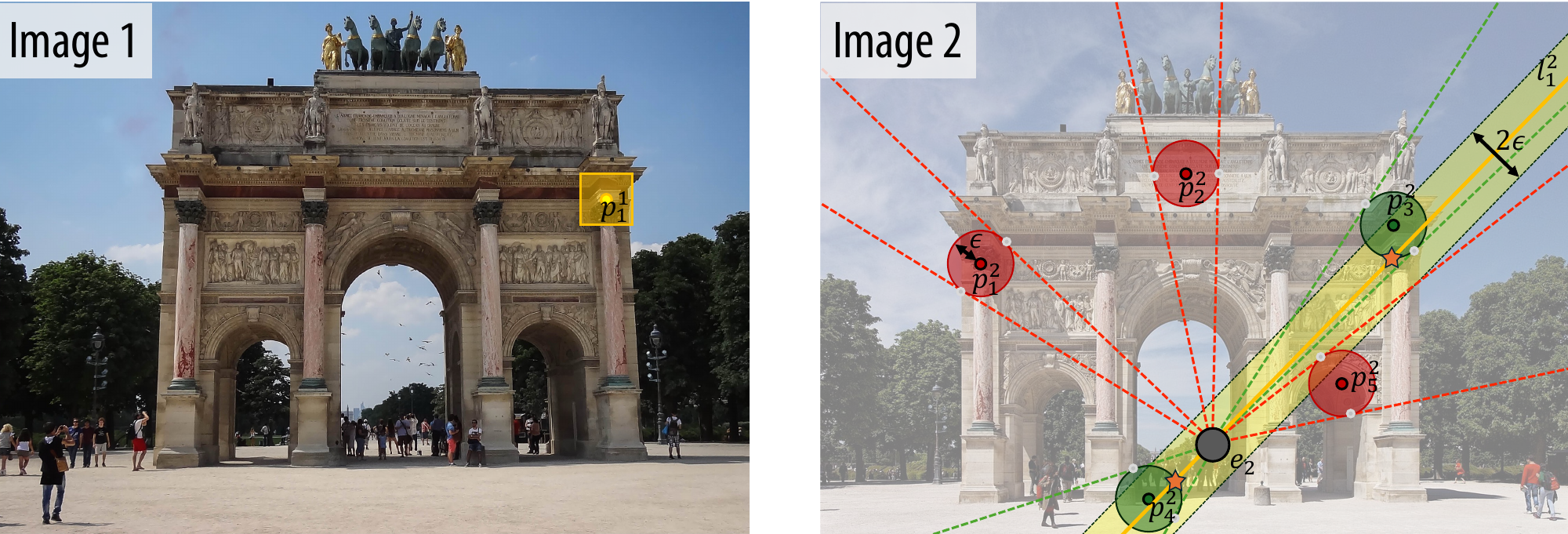}
    \caption{Overview of the guided matching process. Given a query point in image 1 \( \mathbf{p}_1^1 \), its corresponding epipolar line in image 2 \( \mathbf{l}_2^1 \) (yellow) restricts the search space to a narrow envelope of width \(2\epsilon\). Any keypoint lying within this region (e.g., \( \mathbf{p}_4^2 \) and \( \mathbf{p}_3^2 \)) is considered a valid candidate for matching. This can be determined by computing the orthogonal distance from each keypoint to the epipolar line, but such brute-force checks are computationally expensive. Instead, we draw circles of diameter \(2\epsilon\) around each keypoint in image 2 and test whether the epipolar line intersects any of them (orange stars). Interestingly, this geometric test can be reformulated as a fast 1D angular range query: for each circle, the angle interval formed by the two tangents from the epipole defines a valid angular interval, and a candidate is retained if the epipolar line direction lies within it. Green points indicate inliers, red points are outliers. \vspace{-0.5em}} 
    
    \label{fig:teaser_figure}
    \vspace{-0.5em}
\end{figure*}

Matching keypoints between image pairs is an essential component in 3D reconstruction pipelines, such as Structure-from-Motion (SfM)~\cite{triggs1999bundle}, and Simultaneous Localization and Mapping (SLAM)~\cite{mur2017orb}. These tasks are widely used in applications ranging from robotic navigation~\cite{al2024review}, augmented reality~\cite{rameau2022real}, and implicit representation like Gaussian Splatting~\cite{kerbl2024hierarchical}. 
Traditional matching pipelines extract and compare local descriptors to identify tentative correspondences, followed by robust geometric verification to filter outliers~\cite{fischler1981random}. While these approaches can operate without prior geometric knowledge, they become computationally prohibitive and unreliable in challenging scenarios involving repetitive structures, textureless regions, or wide-baseline viewpoints.
When geometric relationships between views are known or can be reliably estimated, they provide powerful constraints that can significantly improve matching efficiency, density, and robustness~\cite{shah2015geometry, barath2021efficient}. The most well-established example is rectified stereo vision, where calibrated cameras reduce correspondence search to efficient 1D scanline matching~\cite{rameau2022mc}.

In the general case of unrectified images with known relative pose or fundamental matrix, one can similarly constrain the search to an \emph{epipolar envelope} --- a narrow band around the epipolar line. This idea, known as epipolar-guided matching, offers the same potential benefits as rectified stereo matching, but is harder to implement efficiently due to the arbitrary orientation and location of epipolar lines in the image.
For that reason, existing methods attempt to approximate the epipolar envelope using angular bins, regular grids, or hashing schemes~\cite{barath2021efficient, shah2015geometry}. While these strategies help reduce complexity, they suffer from two main limitations: they often require additional geometric checks to correct for coarse approximations, and they lack precise control over the envelope width in pixel units. As a result, they may either miss valid candidates or include spurious ones, which weakens both robustness and efficiency.

To cope with these limitations,  we revisit the problem of epipolar-guided matching with a new formulation guaranteeing an exact and fast point retrieval within an epipolar envelope. For this task at hand, rather than computing the pixel-level distance between each keypoint in the second image and the epipolar line, we associate each of these keypoints with a tolerance circle (with a diameter equivalent to the desired epipolar envelope width). Viewed from the epipole, each circle defines an angular interval. A keypoint is retained as a match candidate if its angular interval contains the direction of the epipolar line defined by the query point. This reformulation avoids spatial discretization and enables efficient candidate selection in logarithmic time using a flat segment tree~\cite{de2008computational} (see \figurename~\ref{fig:teaser_figure}).
Beyond avoiding approximation, our method is also faster, as it directly returns only the set of valid keypoints within the epipolar envelope, reducing unnecessary descriptor comparisons or additional geometric checks to eliminate false positive candidates (see \figurename~\ref{fig:approx_compare}). 
Moreover, it offers greater flexibility, allowing per-keypoint tolerance control at pixel-level precision.
Our strategy has been extensively evaluated against existing approaches~\cite{shah2015geometry, barath2021efficient} across a wide range of keypoint densities and epipolar tolerance values, using the ETH3D dataset~\cite{schoeps2017cvpr}. The results highlight better scalability, exactness, and flexibility compared to prior methods. \\
To summarize, the main contributions of this work are as follows:
(i) A novel epipolar-guided matching technique based on fast angular filtering, (ii) Extensive evaluation across varying keypoint densities and tolerance levels, (iii) Open-source implementation.
\section{Related Work}

\begin{figure*}[tb]
    \centering
    \includegraphics[width=0.9\linewidth]{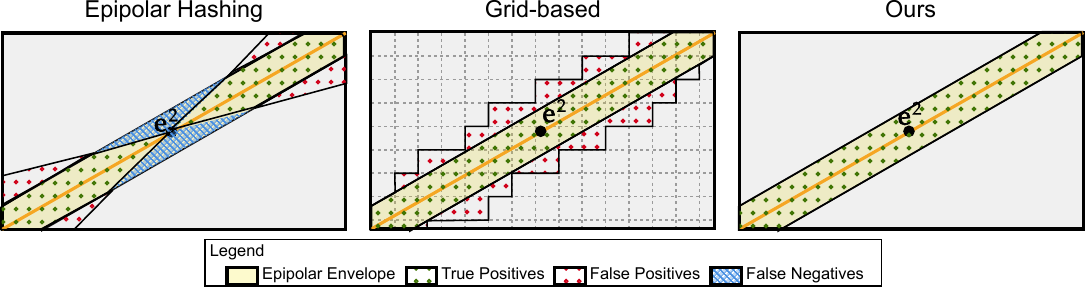}
    \vspace{-0.3em}
    \caption{Comparison of three epipolar correspondence filtering strategies. Unlike our approach, Epipolar Hashing~\cite{barath2021efficient} and Grid-based~\cite{shah2015geometry} methods return matches outside the true epipolar region (false positives) and miss valid matches within it (false negatives). \vspace{-0.5em}}
    \label{fig:approx_compare}
\end{figure*}

Feature matching has long been a fundamental problem in computer vision, with applications in 3D reconstruction~\cite{triggs1999bundle}, localization~\cite{rameau2022real}, and SLAM~\cite{mur2017orb} among many others. As sparse matching scaled to large datasets~\cite{agarwal2011building} and moved into real-time settings~\cite{davison2003real}, the need for faster, more efficient pipelines became critical.
Efforts to accelerate feature extraction led to faster detectors like FAST~\cite{rosten2006machine} and AGAST~\cite{mair2010adaptive}, and more efficient descriptors like SURF~\cite{bay2006surf} and binary alternatives to SIFT~\cite{rublee2011orb}.
Still, the descriptor matching stage remains a bottleneck, especially as the number of keypoints increases.
Initial approaches relied on brute-force matching~\cite{mikolajczyk2005performance}, where each descriptor from one image is compared to all descriptors from the other using a chosen distance metric. This method guarantees that each match corresponds to the true nearest neighbor in descriptor space, but it scales poorly with the number of descriptors. 
To overcome this limitation, Approximate Nearest Neighbor (ANN) methods like FLANN~\cite{muja2009fast}, LSH for binary descriptors~\cite{muja2012fast}, and Product Quantization~\cite{jegou2010product} were introduced to accelerate matching by organizing or compressing descriptors for fast lookup.
Some other works have reformulated the matching problem as a global assignment task using optimal transport~\cite{gold2002graduated}, which minimizes the total cost of pairing features under assignment constraints.
These developments have inspired a new generation of learned matching pipelines, such as SuperGlue~\cite{sarlin2020superglue} and LoFTR~\cite{sun2021loftr} that rely on attention mechanisms and transformers for matching, achieving strong performance at higher computational cost.
Regardless of the specific strategy, most of these pipelines incorporate geometric information only at the final stage, using RANSAC~\cite{fischler1981random} to estimate the fundamental or essential matrix and remove outliers. 
However, in many real-world settings, geometric information between views is already available or can be estimated with reasonable accuracy. This prior can significantly improve matching in terms of robustness, accuracy, and efficiency. For example, SLAM systems often use predicted poses from motion models~\cite{ha20156} or inertial sensors~\cite{qin2018vins} to restrict the correspondence search around projected landmarks~\cite{mur2017orb}. Similarly, in stereo vision, calibrated and rectified cameras reduce the problem to a 1D search along image rows~\cite{chang2018pyramid}, which improves both speed and reliability. These cases show that geometry can do more than just verify matches after the fact, but it can actively guide the matching process itself. All the approaches using prior camera pose information for this purpose can be considered geometric guided matching strategies.
Several prior works have proposed geometry-guided matching strategies conceptually related to ours. One of the earliest geometry-guided approaches is~\cite{shah2015geometry}, which samples points at regular intervals along each epipolar line. For each sample, nearby keypoints are retrieved from a spatial grid, and a local ratio test is applied to reject ambiguous matches in the epipolar envelope. This technique reduces the search space compared to brute-force matching, but still relies on multiple range queries per point. It also introduces several heuristic parameters and requires additional geometric checks, since grid-based retrieval is only an approximation. 
To further reduce complexity, Barath et al.~\cite{barath2021efficient} introduced epipolar hashing, which partitions keypoints into angular bins based on epipolar line orientations for fast candidate lookup. However, this approach assumes angular wedges from the epipole approximate the epipolar envelope, which fails when epipoles lie inside images or near borders. This leads to missed valid matches and inclusion of false candidates, requiring additional Sampson distance checks. Moreover, the tolerance is tied to bin size rather than pixel units, and boundary effects between bins can cause brittle matching. These limitations reduce robustness and adaptability across different camera configurations.
Our method avoids these issues by directly retrieving keypoints within a user-defined pixel tolerance around the epipolar line, without relying on coarse discretization or multi-stage heuristics. It outputs the exact candidate set, so no additional geometric verification is needed. As a result, our approach is faster in practice, even compared to techniques such as epipolar hashing that claim constant-time performance. \figurename~\ref{fig:approx_compare} illustrates the approximation used in existing works and their implications in terms of returned candidates.
Recent learned approaches like Patch2Pix~\cite{zhou2021patch2pix} and Structured Epipolar Matcher~\cite{chang2023structured} incorporate epipolar constraints into neural networks for accuracy-focused refinement. However, these methods prioritize accuracy over efficiency with different computational constraints, making them not directly comparable to our fast geometric filtering approach.

\section{Background and Notations}

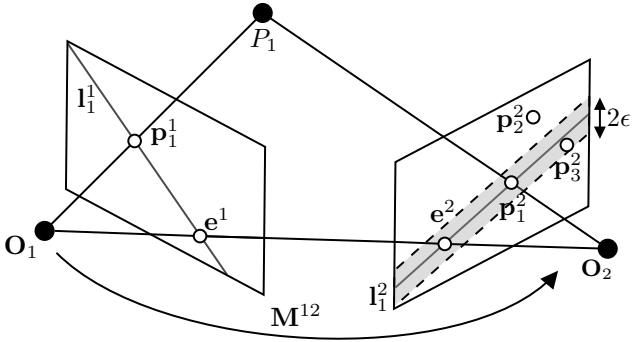
\begin{figure}[tb]
    \centering
    \tikzset{every picture/.style={line width=0.75pt}} 

\begin{tikzpicture}[x=0.75pt,y=0.75pt,yscale=-1,xscale=1]

\draw [color={rgb, 255:red, 74; green, 74; blue, 74 }  ,draw opacity=1 ][fill={rgb, 255:red, 128; green, 128; blue, 128 }  ,fill opacity=1 ]   (239,238.5) -- (158.54,121.68) ;
\draw    (382.41,192.05) -- (431,225.43) ;
\draw [color={rgb, 255:red, 74; green, 74; blue, 74 }  ,draw opacity=1 ][fill={rgb, 255:red, 128; green, 128; blue, 128 }  ,fill opacity=1 ]   (422,157) -- (324,245) ;
\draw  [fill={rgb, 255:red, 155; green, 155; blue, 155 }  ,fill opacity=0.34 ][dash pattern={on 3.75pt off 3.75pt on 3.75pt off 3.75pt}] (421.76,148.67) -- (421.57,167.04) -- (323.21,254.76) -- (323.4,236.39) -- cycle ;
\draw    (192.41,171.05) -- (147,216.43) ;
\draw    (213,150.46) -- (192.41,171.05) ;
\draw    (349,169.5) -- (382.41,192.05) ;
\draw   (258.22,174) -- (257.47,248.67) -- (157.78,195.35) -- (158.54,120.68) -- cycle ;
\draw   (422,130) -- (421.25,204.07) -- (323.21,255.76) -- (323.96,181.69) -- cycle ;
\draw  [draw opacity=0][fill={rgb, 255:red, 0; green, 0; blue, 0 }  ,fill opacity=1 ] (141.94,216.43) .. controls (141.94,213.64) and (144.2,211.37) .. (147,211.37) .. controls (149.8,211.37) and (152.07,213.64) .. (152.07,216.43) .. controls (152.07,219.23) and (149.8,221.5) .. (147,221.5) .. controls (144.2,221.5) and (141.94,219.23) .. (141.94,216.43) -- cycle ;
\draw  [draw opacity=0][fill={rgb, 255:red, 0; green, 0; blue, 0 }  ,fill opacity=1 ] (425.94,225.43) .. controls (425.94,222.64) and (428.2,220.37) .. (431,220.37) .. controls (433.8,220.37) and (436.07,222.64) .. (436.07,225.43) .. controls (436.07,228.23) and (433.8,230.5) .. (431,230.5) .. controls (428.2,230.5) and (425.94,228.23) .. (425.94,225.43) -- cycle ;
\draw  [draw opacity=0][fill={rgb, 255:red, 0; green, 0; blue, 0 }  ,fill opacity=1 ] (251.93,106.45) .. controls (251.93,103.65) and (254.2,101.38) .. (257,101.38) .. controls (259.8,101.38) and (262.07,103.65) .. (262.07,106.45) .. controls (262.07,109.25) and (259.8,111.51) .. (257,111.51) .. controls (254.2,111.51) and (251.93,109.25) .. (251.93,106.45) -- cycle ;
\draw    (257,106.45) -- (349,169.5) ;
\draw  [color={rgb, 255:red, 0; green, 0; blue, 0 }  ,draw opacity=1 ][fill={rgb, 255:red, 255; green, 255; blue, 255 }  ,fill opacity=1 ] (379.23,192.05) .. controls (379.23,190.29) and (380.66,188.87) .. (382.41,188.87) .. controls (384.17,188.87) and (385.59,190.29) .. (385.59,192.05) .. controls (385.59,193.8) and (384.17,195.23) .. (382.41,195.23) .. controls (380.66,195.23) and (379.23,193.8) .. (379.23,192.05) -- cycle ;
\draw    (257,106.45) -- (213,150.46) ;
\draw  [color={rgb, 255:red, 0; green, 0; blue, 0 }  ,draw opacity=1 ][fill={rgb, 255:red, 255; green, 255; blue, 255 }  ,fill opacity=1 ] (189.23,171.05) .. controls (189.23,169.29) and (190.66,167.87) .. (192.41,167.87) .. controls (194.17,167.87) and (195.59,169.29) .. (195.59,171.05) .. controls (195.59,172.8) and (194.17,174.23) .. (192.41,174.23) .. controls (190.66,174.23) and (189.23,172.8) .. (189.23,171.05) -- cycle ;
\draw    (225.41,219.05) -- (152.07,216.43) ;
\draw    (431,225.43) -- (348.82,223) ;
\draw    (324,222) -- (258,220) ;
\draw    (258,220) -- (225.41,219.05) ;
\draw    (258,220) -- (225.41,219.05) ;
\draw    (348.82,223) -- (324,222) ;
\draw  [color={rgb, 255:red, 0; green, 0; blue, 0 }  ,draw opacity=1 ][fill={rgb, 255:red, 255; green, 255; blue, 255 }  ,fill opacity=1 ] (345.64,223) .. controls (345.64,221.24) and (347.06,219.82) .. (348.82,219.82) .. controls (350.58,219.82) and (352,221.24) .. (352,223) .. controls (352,224.76) and (350.58,226.18) .. (348.82,226.18) .. controls (347.06,226.18) and (345.64,224.76) .. (345.64,223) -- cycle ;
\draw  [color={rgb, 255:red, 0; green, 0; blue, 0 }  ,draw opacity=1 ][fill={rgb, 255:red, 255; green, 255; blue, 255 }  ,fill opacity=1 ] (222.23,219.05) .. controls (222.23,217.29) and (223.66,215.87) .. (225.41,215.87) .. controls (227.17,215.87) and (228.59,217.29) .. (228.59,219.05) .. controls (228.59,220.8) and (227.17,222.23) .. (225.41,222.23) .. controls (223.66,222.23) and (222.23,220.8) .. (222.23,219.05) -- cycle ;
\draw [color={rgb, 255:red, 0; green, 0; blue, 0 }  ,draw opacity=1 ]   (427.07,152.11) -- (427.07,167.01) ;
\draw [shift={(427.07,170.01)}, rotate = 270] [fill={rgb, 255:red, 0; green, 0; blue, 0 }  ,fill opacity=1 ][line width=0.08]  [draw opacity=0] (6.25,-3) -- (0,0) -- (6.25,3) -- cycle    ;
\draw [shift={(427.07,149.11)}, rotate = 90] [fill={rgb, 255:red, 0; green, 0; blue, 0 }  ,fill opacity=1 ][line width=0.08]  [draw opacity=0] (6.25,-3) -- (0,0) -- (6.25,3) -- cycle    ;
\draw    (153,227.5) .. controls (204.22,279.21) and (359.25,286.78) .. (404.03,238.73) ;
\draw [shift={(406,236.5)}, rotate = 129.75] [fill={rgb, 255:red, 0; green, 0; blue, 0 }  ][line width=0.08]  [draw opacity=0] (8.93,-4.29) -- (0,0) -- (8.93,4.29) -- cycle    ;
\draw  [color={rgb, 255:red, 0; green, 0; blue, 0 }  ,draw opacity=1 ][fill={rgb, 255:red, 255; green, 255; blue, 255 }  ,fill opacity=1 ] (390.23,159.05) .. controls (390.23,157.29) and (391.66,155.87) .. (393.41,155.87) .. controls (395.17,155.87) and (396.59,157.29) .. (396.59,159.05) .. controls (396.59,160.8) and (395.17,162.23) .. (393.41,162.23) .. controls (391.66,162.23) and (390.23,160.8) .. (390.23,159.05) -- cycle ;
\draw  [color={rgb, 255:red, 0; green, 0; blue, 0 }  ,draw opacity=1 ][fill={rgb, 255:red, 255; green, 255; blue, 255 }  ,fill opacity=1 ] (407.23,173.05) .. controls (407.23,171.29) and (408.66,169.87) .. (410.41,169.87) .. controls (412.17,169.87) and (413.59,171.29) .. (413.59,173.05) .. controls (413.59,174.8) and (412.17,176.23) .. (410.41,176.23) .. controls (408.66,176.23) and (407.23,174.8) .. (407.23,173.05) -- cycle ;

\draw (125,217.4) node [anchor=north west][inner sep=0.75pt]  [font=\normalsize]  {$\mathbf{O}_{1}$};
\draw (416,229.4) node [anchor=north west][inner sep=0.75pt]  [font=\small]  {$\mathbf{O}_{2}$};
\draw (199,158.4) node [anchor=north west][inner sep=0.75pt]  [font=\normalsize]  {$\mathbf{p}_{1}^{1}$};
\draw (248.93,112.85) node [anchor=north west][inner sep=0.75pt]  [font=\normalsize]  {$P_{1}$};
\draw (374.61,196.28) node [anchor=north west][inner sep=0.75pt]  [font=\normalsize]  {$\mathbf{p}_{1}^{2}$};
\draw (429.07,153.51) node [anchor=north west][inner sep=0.75pt]  [font=\normalsize]  {$2\epsilon $};
\draw (309,241.4) node [anchor=north west][inner sep=0.75pt]  [font=\normalsize]  {$\mathbf{l}_{1}^{2}$};
\draw (226,203.4) node [anchor=north west][inner sep=0.75pt]  [font=\normalsize]  {$\mathbf{e}^{1}$};
\draw (340,199.4) node [anchor=north west][inner sep=0.75pt]  [font=\normalsize]  {$\mathbf{e}^{2}$};
\draw (259.47,252.07) node [anchor=north west][inner sep=0.75pt]  [font=\normalsize]  {$\mathbf{M}^{12}$};
\draw (402.23,175.45) node [anchor=north west][inner sep=0.75pt]  [font=\normalsize]  {$\mathbf{p}_{3}^{2}$};
\draw (373.23,151.45) node [anchor=north west][inner sep=0.75pt]  [font=\normalsize]  {$\mathbf{p}_{2}^{2}$};
\draw (162,141.4) node [anchor=north west][inner sep=0.75pt]  [font=\normalsize]  {$\mathbf{l}_{1}^{1}$};

\end{tikzpicture}
    \vspace{-2.5em}
    \caption{Epipolar geometry and envelope-based match filtering. Given a point \( \mathbf{p}_1^1 \) in the first image, its corresponding epipolar line \( \mathbf{l}_1^2 \) is shown in the second image. The grey band illustrates an epipolar envelope of width \( 2\epsilon \), used to retain only candidate matches such as \( \mathbf{p}_1^2 \) and \( \mathbf{p}_3^2 \). \vspace{-0.75em}}
    \label{fig:epipolar_geometry}
    \vspace{-.5em}
\end{figure}

In this work, we consider the classical setup of two calibrated cameras with known extrinsic parameters, --- although knowing only the fundamental matrix is sufficient for epipolar guided matching. As illustrated in \figurename~\ref{fig:epipolar_geometry}, let $\mathbf{p}_i^1 = (x_i^1, y_i^1)^\mathsf{T}$ and $\mathbf{p}_i^2 = (x_i^2, y_i^2)^\mathsf{T}$ denote the projections of the same 3D scene point onto image 1 and image 2, respectively, with homogeneous coordinates $\tilde{\mathbf{p}}_i^1 = (x_i^1, y_i^1, 1)^\mathsf{T}$ and $\tilde{\mathbf{p}}_i^2 = (x_i^2, y_i^2, 1)^\mathsf{T}$.

The relative pose $(\mathbf{R}^{12}, \mathbf{t}^{12})$ induces a geometric relationship between image points, captured by the fundamental matrix $\mathbf{F}^{12}$, which encodes the epipolar geometry:
\begin{equation}
\mathbf{F}^{12} = \left(\mathbf{K}^2\right)^{-\top} [\mathbf{t}^{12}]\times \mathbf{R}^{12} 
\left(\mathbf{K}^1\right)^{-1},
\label{eq:F_mat}
\end{equation}
where $\mathbf{K}^1$ and $\mathbf{K}^2$ are the intrinsic calibration matrices of cameras 1 and 2, respectively, and  \([\mathbf{t}^{12}]_\times\) is the skew-symmetric matrix associated with \(\mathbf{t}^{12}\). 
The fundamental matrix enforces the epipolar constraint:
\begin{equation}
{\left(\tilde{\mathbf{p}}_i^2\right)}^{\top} \mathbf{F}^{12} \tilde{\mathbf{p}}_i^1 = 0,
\end{equation}
which must hold for all true correspondences $\mathbf{p}_i^1 \leftrightarrow \mathbf{p}_i^2$. The epipolar line $\mathbf{l}_i^2 = \mathbf{F}^{12} \tilde{\mathbf{p}}_i^1$ in image 2 defines the locus of potential matches for point $\mathbf{p}_i^1$.
The epipolar geometry also involves the epipoles \(\mathbf{e}^1 = (x_e^1, y_e^1)^\top\) and \(\mathbf{e}^2\), which correspond to the projections of the camera centers onto the opposite image planes. Note that, all epipolar lines \(\mathbf{l}_i^2\) pass through \(\mathbf{e}^2\).

\medskip
\noindent
\textbf{Guided Matching.}  
The goal of the matching process is to find correct correspondences between two sets of keypoints, \(\mathscr{P}^1 = \{\mathbf{p}_1^1, \dots, \mathbf{p}_m^1\}\) in image 1 and \(\mathscr{P}^2 = \{\mathbf{p}_1^2, \dots, \mathbf{p}_n^2\}\) in image 2, based on the similarity between their associated descriptors \(\mathscr{D}^1 = \{\mathbf{D}_1^1, \dots, \mathbf{D}_m^1\}\) and \(\mathscr{D}^2 = \{\mathbf{D}_1^2, \dots, \mathbf{D}_n^2\}\). A common approach is to compare each descriptor \(\mathbf{D}_i^1\) against all descriptors in \(\mathscr{D}^2\), which becomes computationally expensive as \(m\) and \(n\) grow. In the context of guided matching, where the relative pose is known, the search can be restricted geometrically: for each point \(\mathbf{p}_i^1\), only candidate keypoints near its corresponding epipolar line \(\mathbf{l}_i^2\) are considered.
However, due to image noise, discretization, and calibration errors, the epipolar constraint cannot be enforced exactly. We therefore define a pixel-level tolerance $\epsilon$ and introduce the epipolar envelope as the region surrounding $\mathbf{l}_i^2$. Thus, a point $\mathbf{p}_j^2 \in \mathscr{P}^2$ is considered a potential match for $\mathbf{p}_i^1$ if:
\begin{equation}
\mathrm{dist}(\mathbf{l}_i^2, \mathbf{p}_j^2) = \frac{|a_i x_j^2 + b_i y_j^2 + c_i|}{\sqrt{a_i^2 + b_i^2}} \leq \epsilon,
\end{equation}
where $\mathbf{l}_i^2 = (a_i, b_i, c_i)^\top$ is the epipolar line induced by $\mathbf{p}_i^1$.

All such points form the candidate set:
\begin{equation}
\mathscr{C}_i^2 = \left\{ \mathbf{p}_j^2 \in \mathscr{P}^2 \;\middle|\; \mathrm{dist}(\mathbf{l}_i^2, \mathbf{p}_j^2) \leq \epsilon \right\}.
\end{equation}
Each descriptor $\mathbf{D}_i^1$ from image 1 is then compared only against descriptors $\mathbf{D}_j^2$ associated with $\mathscr{C}_i^2$. This guided strategy reduces computational cost and improves robustness in challenging or repetitive environments. 

Nevertheless, computing $\mathscr{C}_i^2$ for all $m$ query points by checking every keypoint in image 2 results in a brute-force complexity of $\mathcal{O}(mn)$. The next section discusses strategies to accelerate this candidate search along epipolar lines.

\section{Methodology}

The core challenge in epipolar-guided matching is to efficiently identify all keypoints in the second image that lie within a pixel tolerance \( \epsilon \) of a given epipolar line. While brute-force approaches suffer from \( \mathcal{O}(mn) \) complexity and approximate methods risk missing valid correspondences, we propose an exact and scalable alternative based on a geometric reformulation of the problem. 
Our approach starts by revisiting the epipolar constraint from an angular perspective, where each keypoint in the second image defines a tolerance region that can be expressed as a 1D angular interval (Sec.~\ref{sec:reformulation}). 
This enables casting candidate search as a fast angular query using a segment tree data structure (Sec.~\ref{sec:angular_query}).
We then show how this geometric filtering integrates with standard descriptor-based strategies, such as Lowe’s ratio~\cite{lowe2004distinctive} test or GMS~\cite{bian2017gms} filtering (Sec.~\ref{sec:desc_matching}). 
Finally, we address implementation considerations, including angular discontinuities and near-epipole cases, to ensure correctness and robustness across configurations (Sec.~\ref{Sec::edge_cases}).

\subsection{Problem Reformulation}
\label{sec:reformulation}

Given a query point $\mathbf{p}_i^1$ in the first image, the corresponding epipolar line $\mathbf{l}_i^2$ in the second image defines a constraint where any valid match must lie within distance $\epsilon$ of this line. Rather than computing the perpendicular distance from each keypoint $\mathbf{p}_j^2$ to $\mathbf{l}_i^2$, we reformulate this constraint using 1D angular intervals viewed from the epipole.
The key geometric insight transforms the question from ``which points are close to the epipolar line?" to ``which epipolar line directions pass close to each point?". To do so, for any keypoint $\mathbf{p}_j^2$ in the second image, we define a tolerance circle of radius $\epsilon$ centered at $\mathbf{p}_j^2$. An epipolar line intersects this circle if and only if the corresponding point $\mathbf{p}_i^1$ represents a potential match under the distance tolerance $\epsilon$.
Viewed from the epipole $\mathbf{e}^2$, each tolerance circle subtends an angular interval. Any epipolar line whose direction falls within this interval will intersect the circle, indicating a valid candidate match. This reformulation transforms the matching problem into a 1D angular range query: given an epipolar line direction $\alpha$, we seek all keypoints whose angular intervals $\Theta$ contain $\alpha$.
\figurename~\ref{fig:interval_query} and Algorithm~\ref{alg:epipolar_matching} summarize the main steps of the angular filtering and range query process.

\subsection{Angular Range Query via Segment Tree}
\label{sec:angular_query}

\begin{figure}[tb]
    \centering
    \resizebox{0.9\linewidth}{!}{%
  \import{figures/}{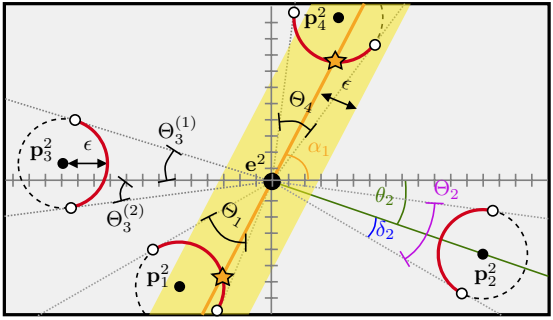}
    }
    \caption{ Illustration of the angular interval query strategy. Given an epipolar line $\mathbf{l}_1^2$ (orange), inlier points lie within the epipolar envelope (yellow). All angular intervals $\Theta_j$ that contain the epipolar line angle $\alpha_1$ correspond to valid match candidates, forming the set $\mathscr{C}_1^2$. In this example, the points $\mathbf{p}_1^2$ and $\mathbf{p}_4^2$ have intervals $\Theta_1$ and $\Theta_4$ that include $\alpha_1$, and are therefore considered inliers. For the point $\mathbf{p}_2^2$, the construction of its interval $\Theta_2$ is illustrated via its central direction $\theta_2$ and angular radius $\delta_2$.   The point $\mathbf{p}_3^2$ exemplifies a wrap-around case, where the interval spans the $0/\pi$ boundary. It is therefore split into two sub-intervals $\Theta_3^{(1)}$ and $\Theta_3^{(2)}$ in the segment tree. \vspace{-0.5em}
}
    \vspace{-0.5em}
    \label{fig:interval_query}
\end{figure}

Each keypoint \( \mathbf{p}_j^2 = (x_j, y_j)^{\top} \) in image 2 is associated with an angular interval that captures its visibility from the epipole \( \mathbf{e}^2 = (x_e, y_e)^{\top} \), under a pixel tolerance \( \epsilon \). This interval corresponds to the directions from the epipole to the two tangents of a circle of radius \( \epsilon \) centered at the keypoint, as illustrated in \figurename~\ref{fig:interval_query}. We first compute the angle from the epipole to the keypoint:
\begin{equation}
\theta_j = \mathrm{atan2}(y_j - y_e,\; x_j - x_e), 
\end{equation}
and the angular radius of the circle seen from the epipole is
\begin{equation}
\delta_j = \arcsin\left(\frac{\epsilon}{\left\| \mathbf{p}_j^2 - \mathbf{e}^2 \right\|} \right).
\end{equation}
The resulting angular interval is defined as \( \Theta_j = [\theta_j - \delta_j,\; \theta_j + \delta_j] \). 
Once computed, all angular intervals \( \Theta_j \) are stored in a flat segment tree structure~\cite{de2008computational}, which supports efficient intersection queries. 

For each query point \( \mathbf{p}_i^1 \) in image 1, we compute the direction \( \alpha_i \) of its corresponding epipolar line \( \mathbf{l}_i^2 \) in image 2. This direction is measured as the angle from the epipole \( \mathbf{e}^2 \) to a reference point \( \mathbf{q}_i \) on the epipolar line, empirically chosen as the point on the line closest to the image center:
\begin{equation}
\alpha_i = \mathrm{atan2}(y_{q_i} - y_e,\; x_{q_i} - x_e).
\end{equation}
Note that while angles span $[0,2\pi)$, epipolar lines are undirected: directions $\alpha$ and $\alpha + \pi$ are equivalent. We therefore work in the reduced domain $[0,\pi)$, mapping all angles via $\alpha' = \alpha \bmod \pi$.

The segment tree is then queried to retrieve all keypoints whose angular intervals $\Theta_j$ contain $\alpha_i$: $\alpha_i \in \Theta_j$.
This filtering step outputs a candidate set \( \mathscr{C}_i^2 \) of matches for \( \mathbf{p}_i^1 \), with complexity \( \mathcal{O}(\log n + k) \), where \( k \) is the number of retrieved candidates. 
Unlike other methods, our approach does not require computing the epipolar distance during querying, leading to faster performance while guaranteeing pixel-level precision and per-point tolerance control, without any approximation.
A complete analysis of time and space complexity for all operations is provided in Table~\ref{tab:complexity}, with the full algorithmic procedure detailed in Algorithm~\ref{alg:epipolar_matching}.
Note that, in practice, $k \gg \log n$, so the total matching complexity becomes $O(\bar{k}m)$, with $\bar{k}$ the average number of points returned per query. Unlike other methods, this time is spent on simple Boolean filtering rather than computing geometric distances.
\begin{algorithm}[htbp]
\caption{Epipolar-Guided Matching }
\label{alg:epipolar_matching}
\begin{algorithmic}[1]
\Require $\mathcal{P}_1,\mathcal{P}_2$, $F_{12}$, tolerance $\epsilon>0$
\Ensure Matches $\mathcal{M}$
\State \textbf{Preprocess:}
\State $e_2 \leftarrow \mathrm{null}(F_{12}^\top)$; \quad $\mathcal{I}\leftarrow\varnothing$
\For{$p_j^2=(x_j,y_j)^\top \in \mathcal{P}_2$}
  \If{$\|p_j^2-e_2\|\le\epsilon$}
    \State add $[0,\pi]$ to $\mathcal{I}$
  \Else
    \State $\theta_j\!\leftarrow\!\mathrm{atan2}(y_j-y_e, x_j-x_e)$
    \State $\delta_j\!\leftarrow\!\arcsin(\epsilon/\|p_j^2-e_2\|)$
    \State constrain $\theta_j$ to $[0, \pi]$ range
    \State add interval $[\theta_j-\delta_j,\;\theta_j+\delta_j]$ to $\mathcal{I}$ \Comment{split at $0/\pi$ if it wraps}
  \EndIf
\EndFor
\State build balanced segment tree $\mathcal{T}$ over $\mathcal{I}$ \Comment{up to $2n$ intervals}
\State \textbf{Match:}
\For{$p_i^1 \in \mathcal{P}_1$}
  \State $l_2^i \leftarrow F_{12}\tilde{p}_1^i$;\; $\alpha_i \leftarrow \mathrm{AngleFromEpipole}(e_2,l_2^i)$
  \State constrain $\alpha_i$ to $[0, \pi]$ range
  \State $\mathcal{C}_i \leftarrow \mathrm{Query}(\mathcal{T},\alpha_i)$ \Comment{expected $O(\log(n)+k)$}
  \State select best by descriptor; validate; add to $\mathcal{M}$
\EndFor
\Return $\mathcal{M}$
\end{algorithmic}
\end{algorithm}
\vspace{-0.5em}

\begin{table}[htbp]
\centering
\caption{Computational Complexity Analysis}
\label{tab:complexity}
\resizebox{\columnwidth}{!}{%
\begin{tabular}{@{}lcc@{}}
\toprule
\textbf{Operation} & \textbf{Time Complexity} & \textbf{Space Complexity} \\
\midrule
Angular interval computation & $O(n)$ & $O(n)$ \\
Segment tree construction & $O(n \log(n)$ & $O(n)$ \\
Single epipolar query & $O(\log(n) + k)$ & $O(1)$ \\
Total matching (all queries) & $O(m(\log(n) + \bar{k}))$ & $O(n)$
\\
\bottomrule
\end{tabular}%
}
\footnotesize
$m$: number of keypoints in image 1, $n$: number of keypoints in image 2, $k$: number of candidates per query, $\bar{k}$: average candidates per query
\vspace{-0.75em}
\end{table}

\subsection{Descriptor Matching and Integration}
\label{sec:desc_matching}
Once geometric filtering has narrowed down the set of potential matches to $\mathcal{C}_i^2 \subset \mathcal{P}^2$, descriptor comparison is restricted to this subset. Specifically, the descriptor $\mathbf{D}_i^1$ of point $\mathbf{p}_i^1$ is matched only against $\mathbf{D}_j^2$ for which $\mathbf{p}_j^2 \in \mathcal{C}_i^2$. This matching strategy reduces computational cost and improves robustness by avoiding comparisons with unrelated descriptors.
Our method is compatible with standard matching strategies, including simple nearest-neighbor matching and the classical Lowe's ratio test~\cite{lowe2004distinctive}. However, guided matching introduces a challenge for traditional ratio-based filtering: the Lowe ratio test relies on comparing the best and second-best matches across the entire descriptor space to detect ambiguous correspondences. In our case, the candidate set $\mathcal{C}_i^2$ may contain relatively few keypoints, potentially making the ratio test less discriminative. To address this limitation, Barath et al.~\cite{barath2021efficient} propose adaptive ratio thresholds that adjust based on the size of the candidate set, though this approach requires careful parameter tuning and may not generalize across different scenes.
As an alternative, GMS~\cite{bian2017gms} is well-suited to guided matching scenarios. By enforcing local motion consistency across image regions without relying on global descriptor distributions, GMS remains effective even when the candidate set is limited.

\subsection{Implementation Details and Special Cases}
\label{Sec::edge_cases}

While the proposed angular filtering method is general and efficient, specific geometric configurations require dedicated handling. 

\vspace{-0.4cm}
\paragraph{Segment Tree Structure}
We use a centered segment tree, adapted to the circular domain. The tree is built recursively, at each node we choose a split angle $c$ --- the median of all midpoints of current intervals. The node stores a bucket of all intervals whose span contains $c$, together with the minimum start and maximum end angles among those intervals. Intervals lying entirely to the left ($end < c$) and right ($start>c$) form the left and right child nodes, respectively. During the querying angle $\alpha$, at each visited node, if $\alpha$ lies between the bucket's min-max bounds, we scan the bucket and return intervals that contain $\alpha$. 

\vspace{-0.4cm}
\paragraph{Discontinuity at $0/\pi$}  
When the angular interval $\Theta_j = [\theta_j - \delta_j,\ \theta_j + \delta_j]$ crosses the $[0, \pi]$ boundary, we split it into two valid sub-intervals:
\[
\Theta_j^{(1)} = [\theta_j - \delta_j, \pi], \quad \Theta_j^{(2)} = [0, \theta_j + \delta_j],
\]
so that both parts can be correctly indexed and queried within the segment tree, this duplication increases the number of intervals stored in the data structure, which may raise the worst-case query complexity to $\mathcal{O}(\log 2n + k)$. However, this has a negligible impact on runtime performance.
\vspace{-0.4cm}
\paragraph{Keypoints near the epipole}  
If the distance between a keypoint $\mathbf{p}_j^2$ and the epipole $\mathbf{e}^2$ is less than $\epsilon$, the angular radius $\delta_j$ becomes undefined. In this case $\mathbf{p}_j^2$ is treated as visible in all directions, with interval $[0,\pi]$, and is always returned during queries since any epipolar line intersects its tolerance region

\section{Experimental Results}
\label{Sec::Experiments}

\begin{figure}[tb]
    \centering
    \includegraphics[width=0.85\linewidth]{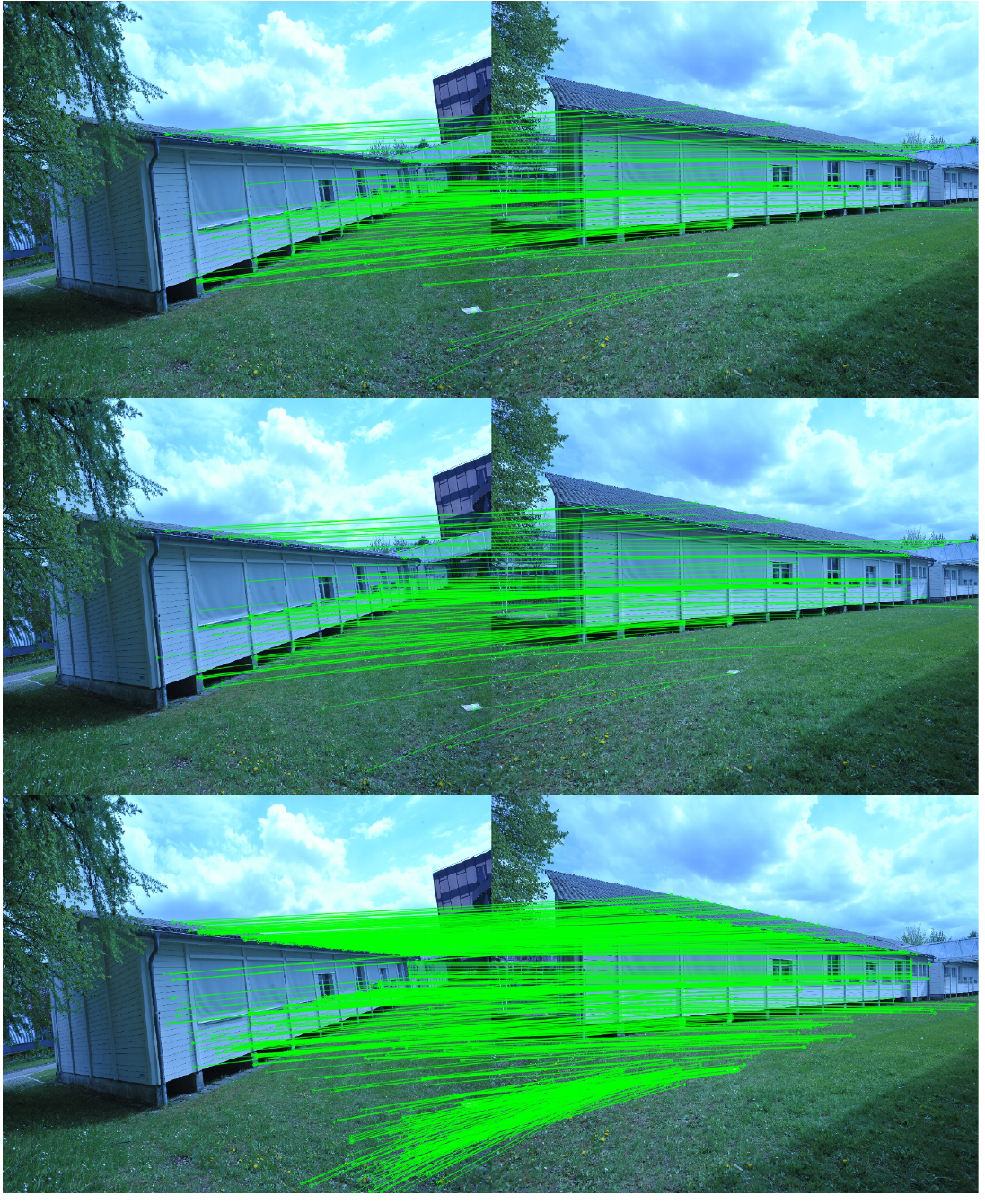}
    \caption{
        Comparison of the number of correctly matched points in a challenging scene with many repetitive structures, such as grass, rooftops, and building facades. (Top) Brute-Force (BF) matching. (Middle) FLANN-based matching. (Bottom) Our epipolar-guided matching. A match is considered correct if the 3D distance between correspondences is below 0.1 meters. \vspace{-0.5em}
    }
    \label{plot:match_num_im}
\end{figure}

\subsection{Dataset and Experimental Setup}
Our method is assessed on the ETH3D dataset~\cite{schoeps2017cvpr}, which consists of 13 high-resolution sequences totaling several hundred images captured in diverse indoor and outdoor environments. 
ETH3D provides high-resolution images up to $6048\times4032$ pixels, dense depth maps, and accurate ground-truth poses, covering scenes with varying levels of texture. Aside from these considerations, ETH3D is also well-suited due to its diverse camera motions. For comparison, autonomous driving datasets such as KITTI~\cite{Geiger2013IJRR} are largely dominated by forward motion, which produces an in-image epipole and would favor our approach. 
\vspace{-0.4cm}
\paragraph{Hardware Setup} All experiments are conducted on a desktop computer running Ubuntu 24.04 LTS with an AMD Ryzen 7 7700X processor and 32GB RAM. To ensure fair comparison, all baseline methods are reimplemented in C++ with identical preprocessing and parallelization.
\vspace{-0.52cm}
\paragraph{Baselines}
We compare against four prior methods: Brute-Force matching, FLANN~\cite{muja2009fast}, Epipolar Hashing~\cite{barath2021efficient}, and Grid-Guided Matching~\cite{shah2015geometry}. All hyperparameters are set as specified in the original works, unless stated otherwise.
All methods use identical descriptor filtering strategies such as Lowe's ratio test ($\tau=0.8$), adaptive Lowe ratio~\cite{barath2021efficient}, or GMS~\cite{bian2017gms} (threshold fixed to 6 neighbors) for a fair comparison. GMS is used by default unless otherwise specified.
Note that for Epipolar Hashing and Grid-Guided Matching, we compute line–point distances instead of the Sampson distance for consistency across methods. Also, to ensure a fair comparison, descriptor-only techniques undergo a final epipolar filtering step that removes matches outside the epipolar envelope, allowing us to incorporate the known geometry. 
\vspace{-0.4cm}
\paragraph{Ground-Truth Correspondences} \noindent
We use SIFT~\cite{lowe2004distinctive} for keypoint detection and description with identical parameters across all methods.  
As ETH3D provides sparse and unevenly distributed triangulated points, we generate denser ground-truth correspondences by associating each detected keypoint with the closest projected 3D point (within 5 pixels) from the provided cleaned scans and camera poses, discarding those without a nearby projection or falling in the occlusion mask.  
A match is considered correct if the 3D distance between the associated points is below 0.1 m (corresponding to less than 1\% of the depth uncertainty even for close objects). Unless stated otherwise, we use the top 50{,}000 keypoints ranked by response.
Only image pairs with at least 500 triangulated keypoints in the dataset ground truth are considered matchable and used for our evaluation. In the case, the final number of matched pairs is lower than 10 (as for ``office" or ``meadow" sequences), we lowered the threshold to 100 points. 
\vspace{-0.4cm}
\paragraph{Evaluation Metrics}
We report the following metrics:  
(i) Total execution time, measured from matching start to final output, excluding feature extraction.
(ii) Candidate generation time and descriptor matching time, reported separately for geometry-guided methods to distinguish the cost of geometric filtering from descriptor comparison.  
(iii) Candidate Recall, the fraction of ground-truth points within the epipolar envelope retrieved by the candidate selection stage (not applicable to methods without geometric filtering).  
(iv) Matching Recall, the fraction of final matches that are correct according to the 3D ground-truth correspondences.

\subsection{Overall Evaluation}

\begin{figure}[tb]
    \centering
    \includegraphics[width=1.1\linewidth]{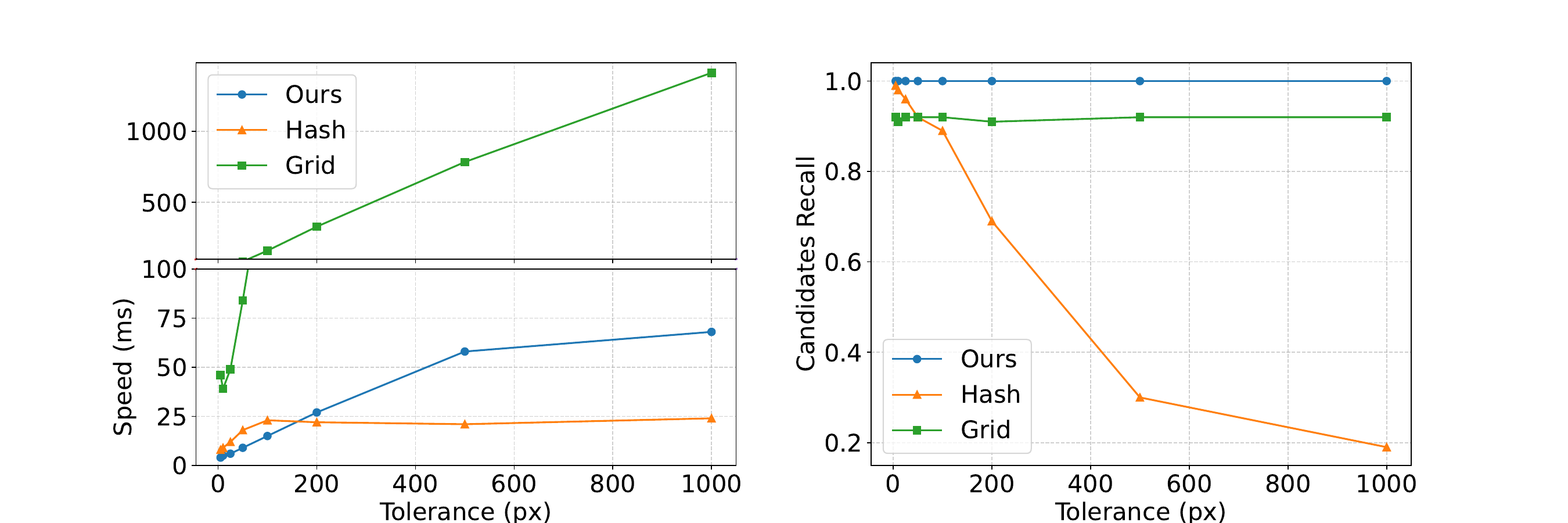}
    \caption{Impact of epipolar tolerance on candidate generation with 25{,}000 keypoints per image. (Left) Candidate generation execution time.  (Right) Candidate set recall. \vspace{-0.5em}}
    \label{plot:tol_vs_speed_precision_court_25k}
\end{figure}

\begin{table*}[t]
\centering
\setlength{\tabcolsep}{2pt}
\resizebox{\textwidth}{!}{
{\small
\begin{tabular}{|l|ccccc|ccc|ccccc|ccccc|}
\hline
 & \multicolumn{5}{c|}{Latency (Cand. Gen. / Desc. Match.) ms.$\downarrow$ } & \multicolumn{3}{c|}{Candidates Recall $\uparrow$ } & \multicolumn{5}{c|}{Matching Recall $\uparrow$} & \multicolumn{5}{c|}{Matches Num.} \\
\cline{2-6}\cline{7-9}\cline{10-14} \cline{15-19}
Sequence   &
 BF & FLANN & Ours & Hash & Grid &
 Ours & Hash & Grid &
 BF & FLANN & Ours & Hash & Grid &
 BF & FLANN & Ours & Hash & Grid \\
\hline
courtyard   &   0/2961 & 0/957 & \bf{34/101} & 76/87 & 145/106 & \bf{1.00} & 0.94 & 0.91 & 0.19 & 0.18 & \bf{0.25} & 0.24  & \bf{0.25} & 4891 &4661 & 28900 & 28917 & 28807
\\
delivery & 0/231 & 0/212 & \bf{2/7} & 11/6 & 22/7  & \bf{1.00} & 0.98 & 0.89 & 0.34 & 0.33 & 0.46 & \bf{0.47} & 0.46 & 3704 & 3537 & 6691 & 6691 & 6677
\\
electro & 0/27 & 0/50 & \bf{2 / 2} &  2 / 3 & 4 / 2 &  \bf{1.00} & 0.85 & 0.91 & 0.21 & 0.19& 0.38 &0.39 & \bf{0.40}  & 977 & 935 & 5880 & 5879 & 5901
\\
facade & 0/2808 & 0/856 &  \bf{35/99} & 73/83 & 362/107 & \bf{1.00} & 0.89 & 0.93 & 0.25 &  0.22 &  \bf{0.35} & \bf{0.35} & 0.34 & 20335 & 19752 & 38282 & 38350 & 38265
\\
kicker & 0/894 & 0/321 & \bf{7/14} & 13/12 & 53/15 & \bf{1.00} & 0.91 & 0.92  & 0.23 & 0.23 & 0.54 & \bf{0.55} & 0.54 & 2589 & 2539 & 10296 & 10103 & 10269
\\
meadow & 0/1426 & 0/262 & \bf{5/16}  & 6/16 & 52 /15 & \bf{1.00} & 0.93 & 0.89 & 0.07 & 0.07 & \bf{0.10} & \bf{0.10} & \bf{0.10} & 1223 &  1177& 8930 & 9082 & 9049
\\
office &  0/29 & 0/50 & \bf{2/2} & \bf{2/2} & 6/2 & \bf{1.00} & 0.94 & 0.93 & 0.11 & 0.10 & \bf{0.21} & \bf{0.21} & \bf{0.21} & 552 & 543 & 3012 & 3077 & 3011
\\
pipes &  0/204 & 0/203 & \bf{2/5} & 8/5 & 20 / 6 & \bf{1.00} & 0.98 & 0.91 & 0.26 & 0.24 & 0.36 & \bf{0.37} & 0.36 & 593 & 524 & 6426 & 6425 & 6438
\\
playground & 0/1323 & 0/553 & \bf{13/34} & 28/31 & 113/38 & \bf{1.00} & 0.89 & 0.93 & 0.22 & 0.18 & \bf{0.34} & 0.33 & \bf{0.34} & 5976 & 5027 & 14321 & 14354 & 14292
\\
relief  & 0/464 & 0/334 &  \bf{3/7}  & 9/7 & 29/8 & \bf{1.00} & 0.96 & 0.91 &  0.24 & 0.20 & \bf{0.48} & \bf{0.48} & \bf{0.48} &1817 & 1726 & 9156 & 9068 & 9090
\\
relief\_2&  0/357 & 0/244 & \bf{3/7} & 11/7 & 33/8 & \bf{1.00} & 0.97 & 0.91 & 0.32 & 0.31 &  \bf{0.57} & \bf{0.57} & \bf{0.57} & 7815 & 7709 & 8089 & 8089 & 8154
\\
terrains  & 0/1695  & 0/691 &  \bf{13/46} & 65/45 & 160/45 & \bf{1.00} & 0.99 & 0.88 & 0.35 & 0.32 & \bf{0.67} & \bf{0.67} & \bf{0.67} & 13333 & 12370 & 23203 & 23194 & 23185
\\
terrace & 0/2051 & 0/744 & \bf{45/146} & 157/ 137 & 335/ 156 & \bf{1.00} & 0.96 & 0.90 & 0.10 & 0.09 & \bf{0.24} & \bf{0.24} & \bf{0.24} &6850 &6306 & 24587 & 24634 & 24690
\\

\hline
\end{tabular}
}
}
\caption{
Median results across all ETH3D~\cite{schoeps2017cvpr} scenes using GMS (Lowe’s ratio results are reported in the supplementary material). Best results are highlighted in \textbf{bold}. For latency, results are compared based on the total time for candidate generation and descriptor matching.  \vspace{-0.5em}
}
\label{tab:main}
\end{table*}


To provide an overview of our method's performance, we evaluate it on all sequences of the ETH3D dataset. 
In this experiment, we assume perfect ground-truth camera poses and intrinsics, which allows us to use an epipolar envelope of $50$ pixels, approximately corresponding to $1\%$ of the image size. 
Qualitative results are available in \figurename~\ref{plot:match_num_im} and Table~\ref{tab:main} reports, for each sequence, the median values over all its image pairs. 
In terms of speed, our method is systematically faster overall. Most of this advantage comes from reduced candidate retrieval time, thanks to the efficiency of our data structure. Interestingly, hashing achieves a lower descriptor matching time, which can be explained by its lower candidate recall: it misses many true candidates, thereby reducing the number of descriptor comparisons. The grid technique is consistently slower due to the large number of queries required to scan the epipolar line. It also shows slightly higher descriptor matching time, as it tends to include false positives that remain for descriptor comparison; these false positives are also reflected in the candidate recall values. Finally, descriptor-only techniques remain significantly slower, and this gap would be even greater with a smaller epipolar envelope.
Regarding matching quality, geometry-guided approaches demonstrate superior recall compared to descriptor-only methods, consistent with prior work~\cite{barath2021efficient,shah2015geometry}. While our approach and the grid-based technique exhibit similar recall, hashing sometimes appears to perform better. Hashing operates with implicitly smaller search regions due to its fixed angular binning constraints, not because it respects the specified pixel tolerance. This creates artificially favorable conditions that break down when poses are imperfect, or epipoles lie near image boundaries, as demonstrated in subsequent experiments in Sec~.~\ref{sec::epi_tol} and Sec~.~\ref{sec::noise_sensi}.

\begin{figure}[tb]
    \centering
    \includegraphics[width=1.0\linewidth]{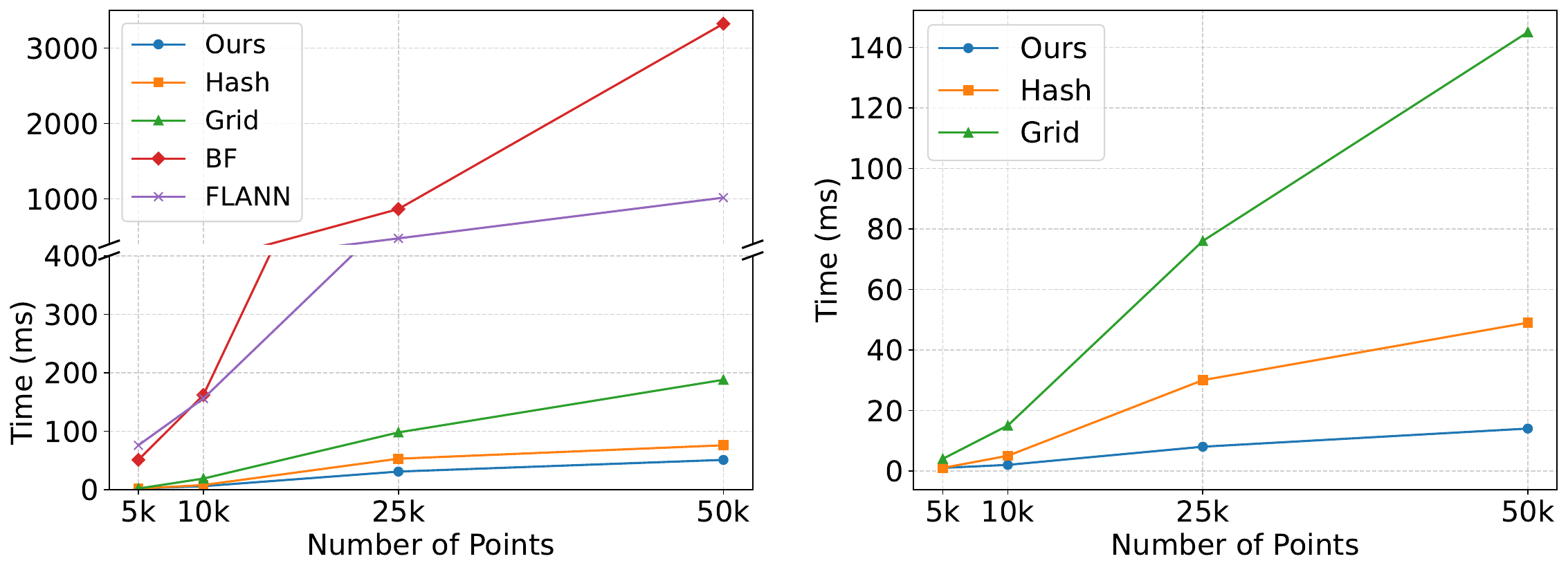}
    \caption{Impact of the number of keypoints on execution latency. 
        (Left) Total matching time. (Right) Candidate set generation time (descriptor matching time is not included). \vspace{-0.5em}}
    \label{plot:speed_test}
\end{figure}

\subsection{Impact of Epipolar Tolerance}
\label{sec::epi_tol}

We evaluated the influence of the epipolar tolerance on matching performance using the entire ``courtyard''sequence, as it yields the highest number of detected keypoints. For consistency, we limited the number of keypoints to 25{,}000 per image. 
We report the \emph{candidate generation execution time} and the \emph{candidate set precision} as the main metrics, as these directly reflect the efficiency of the geometric filtering step and the quality of the resulting candidate set.  
The results are shown in Fig.~\ref{plot:tol_vs_speed_precision_court_25k}.
Note that our approach remains the most effective up to a tolerance of 200~px (roughly 4\% of the image size). Beyond this point, Epipolar Hashing appears faster, but only because its epipolar bin size is fixed.  
As tolerance increases, its bins fail to cover the full requested epipolar envelope, which is clearly visible in the recall curve (the method cannot reliably retrieve all valid points).  
This lack of adaptability can be problematic when the camera pose is approximate and a larger tolerance is needed, as illustrated in Sec .~\ref {sec::noise_sensi}.  
The grid-based technique does not suffer from this limitation but scales significantly worse with tolerance. 


\subsection{Scalability}

We evaluated the scalability of all methods by conducting experiments with 5{,}000, 10{,}000, 25{,}000, and 50{,}000 keypoints on the ``courtyard'' sequence with epipolar threshold of 50 pixels. This scene was chosen because the majority of its images contain over 50{,}000 points due to highly textured surfaces. The results are summarized in Fig.~\ref{plot:speed_test}.
Our method is over 3 times faster than Hashing in candidate generation and about 70\% faster overall, since descriptor comparison dominates the runtime and is comparable across methods.

\subsection{Noise Sensitivity}

\begin{figure}[tb]
    \centering
    \includegraphics[width=0.75\linewidth]{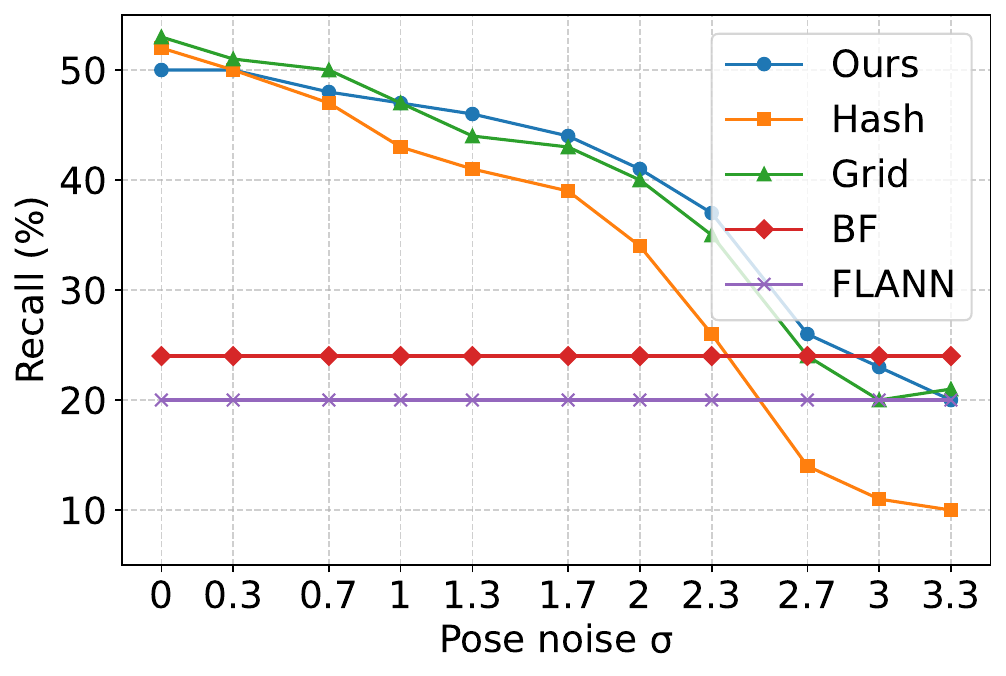}
    \vspace{-1.0em}
    \caption{Matching recall under varying pose noise. Larger noise values correspond to greater perturbations in relative camera pose. }
    \label{plot:noise_lewvels}
    \vspace{-0.75em}
\end{figure}

Geometry-guided methods are inherently sensitive to the accuracy of the fundamental matrix. To evaluate their robustness, we introduced varying levels of noise to the relative pose used in Eq.~\ref{eq:F_mat}. A noise level of 1 corresponds to a randomly perturbed rotation matrix with each axis perturbed within $[-1^\circ, 1^\circ]$, and a translation vector with each component sampled from $[-0.25, 0.25]$ meters — representative of common camera pose estimation errors.
To better simulate roughly calibrated systems, we increased the epipolar threshold to 200 pixels. All experiments were conducted on the ``relief" sequence, which exhibits a wider diversity of camera motions. The results are summarized in Fig.~\ref{plot:noise_lewvels}.
At noise level 0, the Grid and Hash approaches outperform the proposed method, which we attribute to their smaller candidate sets (as discussed in Sec.~\ref{sec::epi_tol}). However, as the noise increases, the performance of the Hash method deteriorates rapidly, indicating limited robustness to pose uncertainty because its fixed bin discretization cannot adapt to the enlarged epipolar envelope, causing it to miss a growing fraction of valid correspondences that are located away from their epipolar lines. In contrast, the Grid-based and Ours maintain stable performance and match the BF baseline at a noise level of 3.3 - approximately corresponding to a translational error of 0.8m and a rotational error of $3^\circ$.

\label{sec::noise_sensi}

\section{Conclusion \& Future Work}

We presented a pixel-accurate epipolar-guided matching method that avoids the discretization errors of existing approaches. By reformulating the problem as a 1D angular interval query from the epipole and solving it with a segment tree, our approach retrieves the exact set of candidates within a user-defined pixel tolerance in logarithmic time. 
This design offers fine-grained control at the keypoint level, eliminates unnecessary descriptor comparisons, and integrates seamlessly into standard matching pipelines.By supporting a per-point matching tolerance, our formulation allows adaptive matching strategies based on available priors, such as pose uncertainty. While not investigated in this paper, this represents a promising direction for future research.
Experiments on ETH3D demonstrate that our method achieves consistent speedups over prior geometric-guided techniques while preserving all valid correspondences. 
While effective in practice, our method has two main limitations: it assumes a finite epipole (epipoles at infinity have not been witnessed during our experiments) and cannot reuse a segment tree for larger tolerances. Since rebuilding adds less than 10\% to execution time, the overhead is minor. Future work will address these cases while also integrating our geometric filtering into learned matching approaches and extending the framework to handle three-view matching.


\section{Acknowledgments}
This work was supported by the Institute of Information \& Communications Technology Planning \& Evaluation(IITP)-Innovative Human Resource Development for Local Intellectualization program grant funded by the Korea government(MSIT)(IITP-2025-RS-2023-00259678).
{
    \small
    \bibliographystyle{ieeenat_fullname}
    \bibliography{main}

@String(CVPR= {IEEE Conf. Comput. Vis. Pattern Recog.})

@String(TOG= {ACM Trans. Graph.})

@String(CVPR  = {CVPR})

@String(TOG   = {ACM TOG})

@article{agarwal2011building,
  title={Building rome in a day},
  author={Agarwal, Sameer and Furukawa, Yasutaka and Snavely, Noah and Simon, Ian and Curless, Brian and Seitz, Steven M and Szeliski, Richard},
  journal={Communications of the ACM},
  volume={54},
  number={10},
  pages={105--112},
  year={2011},
  publisher={ACM New York, NY, USA}
}

@inproceedings{davison2003real,
  title={Real-time simultaneous localisation and mapping with a single camera},
  author={Davison},
  booktitle={Proceedings Ninth IEEE International Conference on Computer Vision},
  pages={1403--1410},
  year={2003},
  organization={IEEE}
}

@article{mur2017orb,
  title={Orb-slam2: An open-source slam system for monocular, stereo, and rgb-d cameras},
  author={Mur-Artal, Raul and Tard{\'o}s, Juan D},
  journal={IEEE transactions on robotics},
  volume={33},
  number={5},
  pages={1255--1262},
  year={2017},
  publisher={IEEE}
}

@inproceedings{triggs1999bundle,
  title={Bundle adjustment—a modern synthesis},
  author={Triggs, Bill and McLauchlan, Philip F and Hartley, Richard I and Fitzgibbon, Andrew W},
  booktitle={International workshop on vision algorithms},
  pages={298--372},
  year={1999},
  organization={Springer}
}

@article{rameau2022real,
  title={Real-time multi-car localization and see-through system},
  author={Rameau, Francois and Bailo, Oleksandr and Park, Jinsun and Joo, Kyungdon and Kweon, In So},
  journal={International Journal of Computer Vision},
  volume={130},
  number={2},
  pages={384--404},
  year={2022},
  publisher={Springer}
}

@article{kerbl2024hierarchical,
  title={A hierarchical 3d gaussian representation for real-time rendering of very large datasets},
  author={Kerbl, Bernhard and Meuleman, Andreas and Kopanas, Georgios and Wimmer, Michael and Lanvin, Alexandre and Drettakis, George},
  journal={ACM Transactions on Graphics (TOG)},
  volume={43},
  number={4},
  pages={1--15},
  year={2024},
  publisher={ACM New York, NY, USA}
}

@article{al2024review,
  title={A review of visual SLAM for robotics: Evolution, properties, and future applications},
  author={Al-Tawil, Basheer and Hempel, Thorsten and Abdelrahman, Ahmed and Al-Hamadi, Ayoub},
  journal={Frontiers in Robotics and AI},
  volume={11},
  pages={1347985},
  year={2024},
  publisher={Frontiers Media SA}
}

@article{rameau2022mc,
  title={MC-Calib: A generic and robust calibration toolbox for multi-camera systems},
  author={Rameau, Francois and Park, Jinsun and Bailo, Oleksandr and Kweon, In So},
  journal={Computer Vision and Image Understanding},
  volume={217},
  pages={103353},
  year={2022},
  publisher={Elsevier}
}

@inproceedings{rosten2006machine, 
  title={Machine learning for high-speed corner detection},
  author={Rosten, Edward and Drummond, Tom},
  booktitle={European conference on computer vision},
  pages={430--443},
  year={2006},
  organization={Springer}
}

@inproceedings{mair2010adaptive,
  title={Adaptive and generic corner detection based on the accelerated segment test},
  author={Mair, Elmar and Hager, Gregory D and Burschka, Darius and Suppa, Michael and Hirzinger, Gerhard},
  booktitle={European conference on Computer vision},
  pages={183--196},
  year={2010},
  organization={Springer}
}

@article{lowe2004distinctive,
  title={Distinctive image features from scale-invariant keypoints},
  author={Lowe, David G},
  journal={International journal of computer vision},
  volume={60},
  number={2},
  pages={91--110},
  year={2004},
  publisher={Springer}
}

@inproceedings{bay2006surf,
  title={Surf: Speeded up robust features},
  author={Bay, Herbert and Tuytelaars, Tinne and Van Gool, Luc},
  booktitle={European conference on computer vision},
  pages={404--417},
  year={2006},
  organization={Springer}
}

@inproceedings{rublee2011orb,
  title={ORB: An efficient alternative to SIFT or SURF},
  author={Rublee, Ethan and Rabaud, Vincent and Konolige, Kurt and Bradski, Gary},
  booktitle={2011 International conference on computer vision},
  pages={2564--2571},
  year={2011},
  organization={Ieee}
}

@article{mikolajczyk2005performance,
  title={A performance evaluation of local descriptors},
  author={Mikolajczyk, Krystian and Schmid, Cordelia},
  journal={IEEE transactions on pattern analysis and machine intelligence},
  volume={27},
  number={10},
  pages={1615--1630},
  year={2005},
  publisher={IEEE}
}

@article{muja2009fast,
  title={Fast approximate nearest neighbors with automatic algorithm configuration.},
  author={Muja, Marius and Lowe, David G},
  journal={VISAPP (1)},
  volume={2},
  number={331-340},
  pages={2},
  year={2009}
}

@inproceedings{muja2012fast,
  title={Fast matching of binary features},
  author={Muja, Marius and Lowe, David G},
  booktitle={2012 Ninth conference on computer and robot vision},
  pages={404--410},
  year={2012},
  organization={IEEE}
}

@article{jegou2010product,
  title={Product quantization for nearest neighbor search},
  author={Jegou, Herve and Douze, Matthijs and Schmid, Cordelia},
  journal={IEEE transactions on pattern analysis and machine intelligence},
  volume={33},
  number={1},
  pages={117--128},
  year={2010},
  publisher={IEEE}
}

@inproceedings{bian2017gms,
  title={Gms: Grid-based motion statistics for fast, ultra-robust feature correspondence},
  author={Bian, JiaWang and Lin, Wen-Yan and Matsushita, Yasuyuki and Yeung, Sai-Kit and Nguyen, Tan-Dat and Cheng, Ming-Ming},
  booktitle={Proceedings of the IEEE conference on computer vision and pattern recognition},
  pages={4181--4190},
  year={2017}
}

@article{gold2002graduated,
  title={A graduated assignment algorithm for graph matching},
  author={Gold, Steven and Rangarajan, Anand},
  journal={IEEE Transactions on pattern analysis and machine intelligence},
  volume={18},
  number={4},
  pages={377--388},
  year={2002},
  publisher={IEEE}
}

@inproceedings{sarlin2020superglue,
  title={Superglue: Learning feature matching with graph neural networks},
  author={Sarlin, Paul-Edouard and DeTone, Daniel and Malisiewicz, Tomasz and Rabinovich, Andrew},
  booktitle={Proceedings of the IEEE/CVF conference on computer vision and pattern recognition},
  pages={4938--4947},
  year={2020}
}

@inproceedings{sun2021loftr,
  title={LoFTR: Detector-free local feature matching with transformers},
  author={Sun, Jiaming and Shen, Zehong and Wang, Yuang and Bao, Hujun and Zhou, Xiaowei},
  booktitle={Proceedings of the IEEE/CVF conference on computer vision and pattern recognition},
  pages={8922--8931},
  year={2021}
}

@inproceedings{barath2021efficient,
  title={Efficient initial pose-graph generation for global sfm},
  author={Barath, Daniel and Mishkin, Dmytro and Eichhardt, Ivan and Shipachev, Ilia and Matas, Jiri},
  booktitle={Proceedings of the IEEE/CVF Conference on Computer Vision and Pattern Recognition},
  pages={14546--14555},
  year={2021}
}

@inproceedings{shah2015geometry,
  title={Geometry-aware feature matching for structure from motion applications},
  author={Shah, Rajvi and Srivastava, Vanshika and Narayanan, PJ},
  booktitle={2015 IEEE Winter Conference on Applications of Computer Vision},
  pages={278--285},
  year={2015},
  organization={IEEE}
}

@inproceedings{zhou2021patch2pix,
  title={Patch2pix: Epipolar-guided pixel-level correspondences},
  author={Zhou, Qunjie and Sattler, Torsten and Leal-Taixe, Laura},
  booktitle={Proceedings of the IEEE/CVF conference on computer vision and pattern recognition},
  pages={4669--4678},
  year={2021}
}

@inproceedings{chang2023structured,
  title={Structured epipolar matcher for local feature matching},
  author={Chang, Jiahao and Yu, Jiahuan and Zhang, Tianzhu},
  booktitle={Proceedings of the IEEE/CVF conference on computer vision and pattern recognition},
  pages={6177--6186},
  year={2023}
}

@article{fischler1981random,
  title={Random sample consensus: a paradigm for model fitting with applications to image analysis and automated cartography},
  author={Fischler, Martin A and Bolles, Robert C},
  journal={Communications of the ACM},
  volume={24},
  number={6},
  pages={381--395},
  year={1981},
  publisher={ACM New York, NY, USA}
}

@inproceedings{chang2018pyramid,
  title={Pyramid stereo matching network},
  author={Chang, Jia-Ren and Chen, Yong-Sheng},
  booktitle={Proceedings of the IEEE conference on computer vision and pattern recognition},
  pages={5410--5418},
  year={2018}
}

@article{qin2018vins,
  title={Vins-mono: A robust and versatile monocular visual-inertial state estimator},
  author={Qin, Tong and Li, Peiliang and Shen, Shaojie},
  journal={IEEE transactions on robotics},
  volume={34},
  number={4},
  pages={1004--1020},
  year={2018},
  publisher={IEEE}
}

@incollection{ha20156,
  title={6-DOF direct homography tracking with extended Kalman filter},
  author={Ha, Hyowon and Rameau, Fran{\c{c}}ois and Kweon, In So},
  booktitle={Image and Video Technology},
  pages={447--460},
  year={2015},
  publisher={Springer}
}

@book{de2008computational,
  title={Computational geometry: algorithms and applications},
  author={De Berg, Mark and Cheong, Otfried and Van Kreveld, Marc and Overmars, Mark},
  year={2008},
  publisher={Springer}
}

@inproceedings{schoeps2017cvpr,
  author = {Thomas Sch\"ops and Johannes L. Sch\"onberger and Silvano Galliani and Torsten Sattler and Konrad Schindler and Marc Pollefeys and Andreas Geiger},
  title = {A Multi-View Stereo Benchmark with High-Resolution Images and Multi-Camera Videos},
  booktitle = {Conference on Computer Vision and Pattern Recognition (CVPR)},
  year = {2017}
}

@article{Geiger2013IJRR,
  author = {Andreas Geiger and Philip Lenz and Christoph Stiller and Raquel Urtasun},
  title = {Vision meets Robotics: The KITTI Dataset},
  journal = {International Journal of Robotics Research (IJRR)},
  year = {2013}
}
}
\clearpage
\setcounter{page}{1}
\maketitlesupplementary

\section{Lowe's Ration Test Results}
The results of Lowe's ratio test are reported in Table~\ref{tab:lowe}. As shown, the ratio test requires roughly the same computation time, but produces more candidate matches since each descriptor is compared only within the epipolar envelope rather than across the entire image. This naturally improves matching recall, but it also increases the number of false positives.
  
\section{Segment Tree Construction Latency}
\begin{figure}[h]
    \centering
    \includegraphics[width=0.9\linewidth]{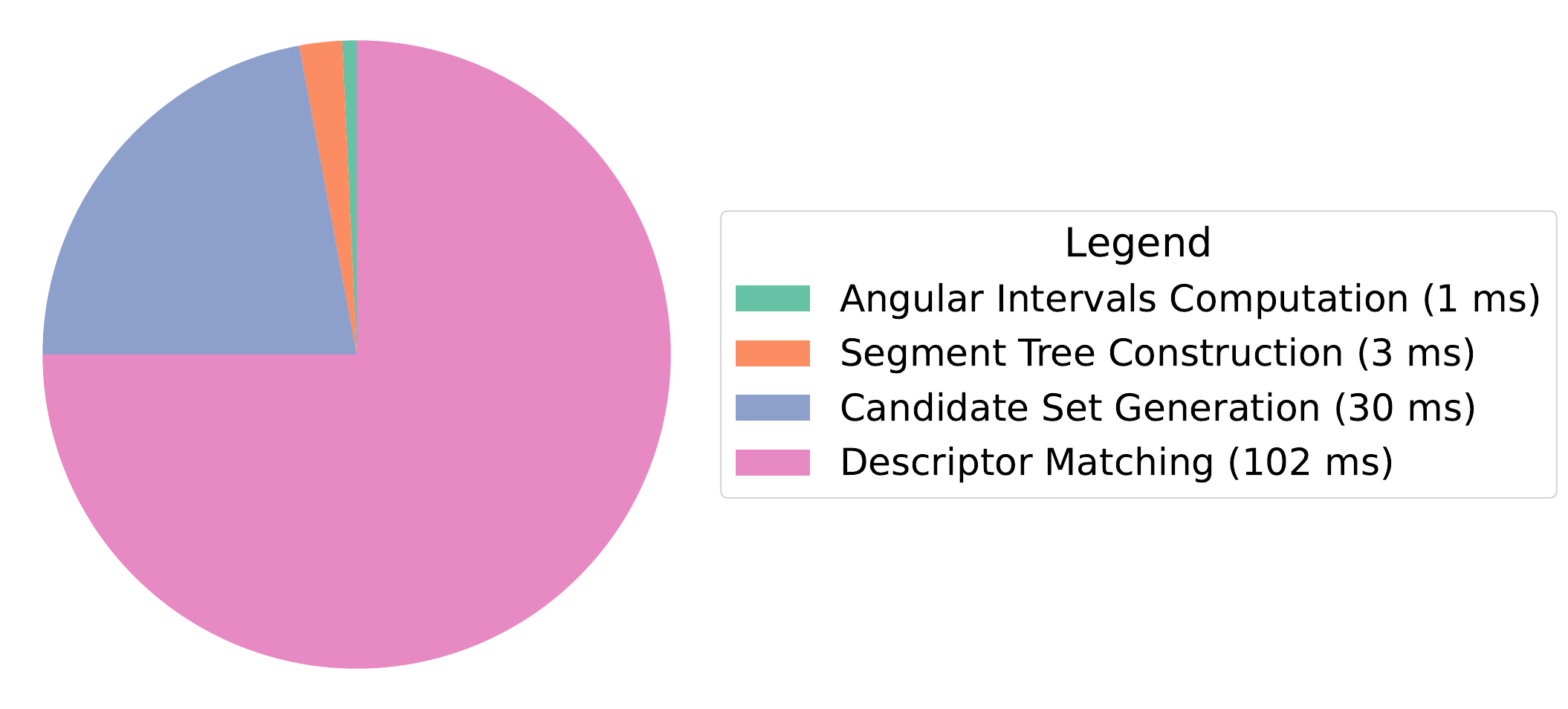}
    \caption{Point Matching Latency Breakdown.}
    \label{fig:latency_break}
\end{figure}

To better understand the computational overhead of our data structure, we measured the latency of segment tree construction on the ``courtyard'' sequence, which contains one of the highest numbers of keypoints per image in the ETH3D dataset. The total runtime of this stage was approximately 34 ms. Out of this, around 1 ms was spent on computing the angular intervals for all keypoints, 3 ms on the actual construction of the balanced segment tree, and the remaining 30 ms on querying the candidate points against the tree. In contrast, the descriptor matching takes 102 ms. This shows that the cost of building the data structure itself is negligible compared to the subsequent query operations, as illustrated in \figurename~\ref{fig:latency_break}. 

\begin{figure*}[htbp]
    \centering
    \includegraphics[width=0.9\linewidth]{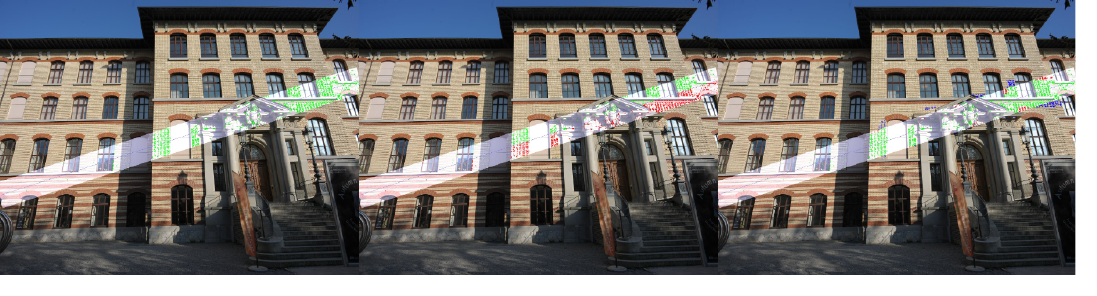}
    \caption{Comparison of three epipolar correspondence filtering strategies. (Left) Ours. (Middle) Epipolar Hashing. (Right) Grid-based. Green points represent true positive candidates, red - false negatives, and blue - false positives.}
    \label{fig:3_env_1}
\end{figure*}

\begin{figure*}[htbp]
    \centering
    \includegraphics[width=0.9\linewidth]{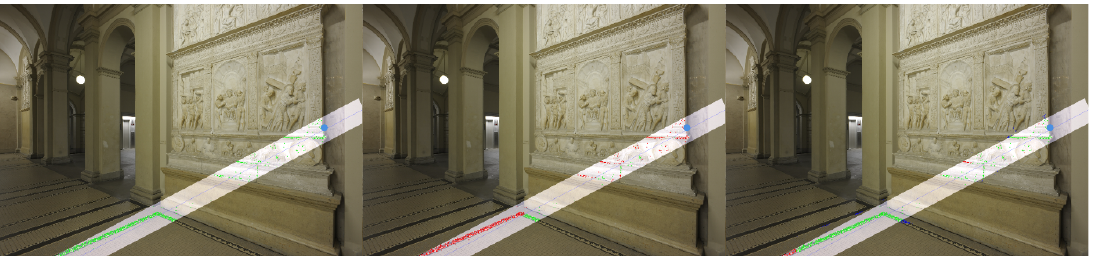}
    \caption{Comparison of three epipolar correspondence filtering strategies when the epipole lies inside the image. (Left) Ours. (Middle) Epipolar Hashing. (Right) Grid-based. Green points represent true positive candidates, red - false negatives, blue - false positives. The large light blue point indicates the epipole.}
    \label{fig:3_env_2}
\end{figure*}

\section{Candidate Set Generation}
To better demonstrate what candidate sets are produced by different methods and how they influence matching, we visualize them in Fig.~\ref{fig:3_env_1} and Fig.~\ref{fig:3_env_2}. As can be seen from the images, the Epipolar Hashing approach often fails to include many valid candidates. Moreover, as shown in Fig.~\ref{fig:3_env_1}, points located near bin boundaries can be excluded, and even small perturbations in the fundamental matrix can lead to missed valid matches.
Additionally, when the epipole is located within the image, as in Fig.~\ref{fig:3_env_2}, many valid candidates are omitted because they lie too close to the epipole and are unevenly distributed during binning. The grid-based approach preserves most valid points but tends to produce false positives. In contrast, our method consistently returns the complete set of valid candidates.

\begin{table*}[t]
\centering
\setlength{\tabcolsep}{2pt}
\resizebox{\textwidth}{!}{
{\small
\begin{tabular}{|l|ccccc|ccc|ccccc|ccccc|}
\hline
 & \multicolumn{5}{c|}{Latency (Cand. Gen. / Desc. Match.) ms.$\downarrow$ } & \multicolumn{3}{c|}{Candidates Recall $\uparrow$ } & \multicolumn{5}{c|}{Matching Recall $\uparrow$} & \multicolumn{5}{c|}{Matches Num.} \\
\cline{2-6}\cline{7-9}\cline{10-14} \cline{15-19}
Sequence   &
 BF & FLANN & Ours & Hash & Grid &
 Ours & Hash & Grid &
 BF & FLANN & Ours & Hash & Grid &
 BF & FLANN & Ours & Hash & Grid \\
\hline
courtyard   &   0/2961 & 0/957 & 34/102 & 76/87 & 145/106 & 1.00 & 0.94 & 0.91 & 0.19 & 0.18 & 0.29 & 0.30  & 0.29 & 4891 &4661 & 37346 & 37346 & 37346
\\
delivery & 0/231 & 0/212 & 2/8 & 11/8 & 22/8  & 1.00 & 0.98 & 0.89 & 0.34 & 0.33 & 0.57 &0.57 & 0.57 & 3704 & 3537 & 10858 & 10858 & 10858
\\
electro & 0/27 & 0/50 & 2 / 2 &  2 / 2 & 4 / 2 &  1.00 & 0.85 & 0.91 & 0.21 & 0.19& 0.67 & 0.66 & 0.67  & 977 & 935 & 8534 & 8578 & 8531
\\
facade & 0/2808 & 0/856 &  35/104 & 73/89 & 362/90 & 1.00 & 0.89 & 0.93 & 0.25 &  0.22 &  0.39 & 0.39 & 0.39 & 20335 & 19752 & 45344 & 45157 & 45126
\\
kicker & 0/894 & 0/321 & 7/15 & 13/13 & 53/14 & 1.00 & 0.91 & 0.92  & 0.23 & 0.23 & 0.63 & 0.63 & 0.63 & 2589 & 2539 & 14441 & 14438 & 14435
\\
meadow & 0/1426 & 0/262 & 5/13  & 6/13 & 52/13 & 1.00 & 0.93 & 0.89 & 0.07 & 0.07 & 0.15 & 0.15 & 0.15 & 1223 &  1177& 11815 & 11815 & 11815
\\
office &  0/29 & 0/50 & 2/1 & 2/1 & 6/1 & 1.00 & 0.94 & 0.93 & 0.11 & 0.10 & 0.32 & 0.32 & 0.32 & 552 & 543 & 4836 & 4836 & 4836
\\
pipes &  0/204 & 0/203 & 2/4 & 8/4 & 20 /4 & 1.00 & 0.98 & 0.91 & 0.26 & 0.24 & 0.43 & 0.44 & 0.44 & 593 & 524 & 9631 & 9631 & 9601
\\
playground & 0/1323 & 0/553 & 13/53 & 28/32 & 113/35 & 1.00 & 0.89 & 0.93 & 0.22 & 0.18 & 0.42 & 0.42 & 0.42 & 5976 & 5027 & 21880 & 21873 & 21880
\\
relief  & 0/464 & 0/334 &  3/8  & 9/8 & 29/8 & 1.00 & 0.96 & 0.91 &  0.24 & 0.20 & 0.60 & 0.60 & 0.60 &1817 & 1726 & 14566 & 14556 & 14512
\\
relief\_2&  0/357 & 0/244 & 3/8 & 11/8 & 33/8 & 1.00 & 0.97 & 0.91 & 0.32 & 0.31 &  0.68 & 0.69 & 0.69 & 7815 & 7709 & 12559 & 12543 & 12543
\\
terrains  & 0/1695  & 0/691 &  13/50 & 65/50 & 160/41 & 1.00 & 0.99 & 0.88 & 0.35 & 0.32 & 0.78 & 0.78 & 0.78 & 13333 & 12370 & 26587 & 26587 & 26587
\\
terrace & 0/2051 & 0/744 & 45/152 & 157/134 & 335/139 & 1.00 & 0.96 & 0.90 & 0.10 & 0.09 & 0.27 & 0.27 & 0.28 &6850 &6306 & 37584 & 37584 & 37584
\\

\hline
\end{tabular}
}
}
\caption{
Median results across all ETH3D~\cite{schoeps2017cvpr} scenes using Lowe’s ratio.
}
\label{tab:lowe}
\end{table*}


\end{document}